\documentclass[runningheads]{llncs}

\usepackage{graphics} 
\usepackage{subfigure}
\usepackage{epsfig} 
\usepackage{amsmath} 
\usepackage{amssymb}  
\usepackage{balance}
\usepackage{xcolor}
\usepackage{url}
\usepackage[mathscr]{euscript}
\usepackage{soul}
\usepackage{hyperref}

\newcommand\set[1]{\mathscr{#1}}
\newcommand\rosmeery{\textsc{RoSmEEry}}
\newcommand\equationname{Eq.}
\newcommand\sectionname{Section}
\newcommand{\abs}[1]{\left\lvert #1 \right\rvert}

\begin{document}
\title{RoSmEEry: Robotic Simulated Environment for Evaluation and Benchmarking of Semantic Mapping Algorithms}
%
%
\author{
Sara Kaszuba\inst{1}
\and Sandeep Reddy Sabbella\inst{1}
\and Vincenzo Suriani\inst{1}\orcidID{0000-0003-1199-8358}
\and \\ Francesco Riccio\inst{1}\orcidID{0000-0002-9112-8143}
\and Daniele Nardi\inst{1}\orcidID{0000-0001-6606-200X}
}
\authorrunning{Kaszuba et al.}
%
\titlerunning{Robotic Simulated Environment for Semantic Mapping Benchmarking}
\institute{The Department of Computer, Control, and Management Engineering Antonio Ruberti (DIAG) of Sapienza University of Rome, via Ariosto 25, Rome, Italy. 
    \email{\{lastname\}@diag.uniroma1.it.}
}

\maketitle              

\begin{abstract}
Human-robot interaction requires a common understanding of the operational environment, which can be provided by a representation that blends geometric and symbolic knowledge: a \textit{semantic map}. Through a semantic map the robot can interpret user commands by grounding them to its sensory observations. Semantic mapping is the process that builds such a representation. Despite being fundamental to enable cognition and high-level reasoning in robotics, semantic mapping is a challenging task due to generalization to different scenarios and sensory data types. In fact, it is difficult to obtain a rich and accurate semantic map of the environment and of the objects therein. Moreover, to date, there are no frameworks that allow for a comparison of the performance in building semantic maps for a given environment. To tackle these issues we design \rosmeery, a novel framework based on the \textit{Gazebo} simulator, where we introduce an accessible and ready-to-use methodology for a systematic evaluation of semantic mapping algorithms.  We release our framework, as an open-source package, with multiple simulation environments with the aim to provide a general set-up to quantitatively measure the performances in acquiring semantic knowledge about the environment.

\end{abstract}
\section{INTRODUCTION}
\label{sec:intro}
Semantic mapping is fundamental to bridge human semantic knowledge to robot sensory observations. Nevertheless, building a comprehensive map of the environment for a robot -- spanning from raw sensory observation to high-level semantics~\cite{Pronobis2017} -- is an extremely difficult task~\cite{taketomi2017visual}. Moreover, in literature there is not a valid methodology to evaluate the performance of a semantic mapping system nor a ready-to-use benchmarking framework. As a consequence, proposed approaches are built with strong assumptions on the domain of application and they usually improve on results achieved with specific environment settings -- being hardly reproducible. To tackle this issue, in this paper, we introduce a \textbf{Ro}botic \textbf{S}i\textbf{m}ulated \textbf{E}nvironment for \textbf{E}valuation and Benchma\textbf{r}king of Semantic Mapping algorithms, that we refer to as \rosmeery{}. Such a novel simulation environment allows to systematically evaluate semantic mapping algorithms by relying on the semantic map formalization introduced in~\cite{Capobianco2015}.  Our proposal is implemented through a simulation environment developed in \textit{Gazebo}. Noticeably, our approach not only evaluates the geometrical accuracy of the map, but also the amount and quality of the knowledge represented in the semantic map. Furthermore, our goal (and scope for this contribution) is to evaluate the process of gathering semantic knowledge of and environment, whilst the evaluation of how such a knowledge is exploited, is deferred to future work.

To demonstrate how \rosmeery{} can be used to evaluate, and benchmark, semantic mapping algorithms, we provide different exploration policies that serve as baselines. Nevertheless, such baselines represent a proxy to show-case the evaluation framework features and they have to be replaced, or compared, with the semantic mapping algorithm under examination. Specifically, we implemented a standard frontier-based exploration strategy where the robot visits always the biggest frontier in the map; a random exploration policy where the robot randomly roams the environment; and a human-driven exploration policy where a user is tasked to explore an environment with the goal to locate as many semantic objects as possible. Moreover, in order to have a comparable test in the human-driven exploration, users navigate the environment by looking at the same images that a robot could perceive through its sensors.

However, despite the exploration policy used, the robotic agent within the \rosmeery{} framework is equipped with a spatial knowledge database that keeps track of the semantic knowledge observed during the benchmark, and matches the gathered information against the groundtruth provided by the simulator. Then, the quality of the semantic mapping session is quantitatively evaluated in accordance with a set of desired metrics. It is fundamental to understand that such metrics can be defined by the user and can be dynamically adjusted at the beginning of each session to accommodate the desired benchmark configuration.
\begin{figure}[!t]
    \centering
    \includegraphics[width=0.92\columnwidth]{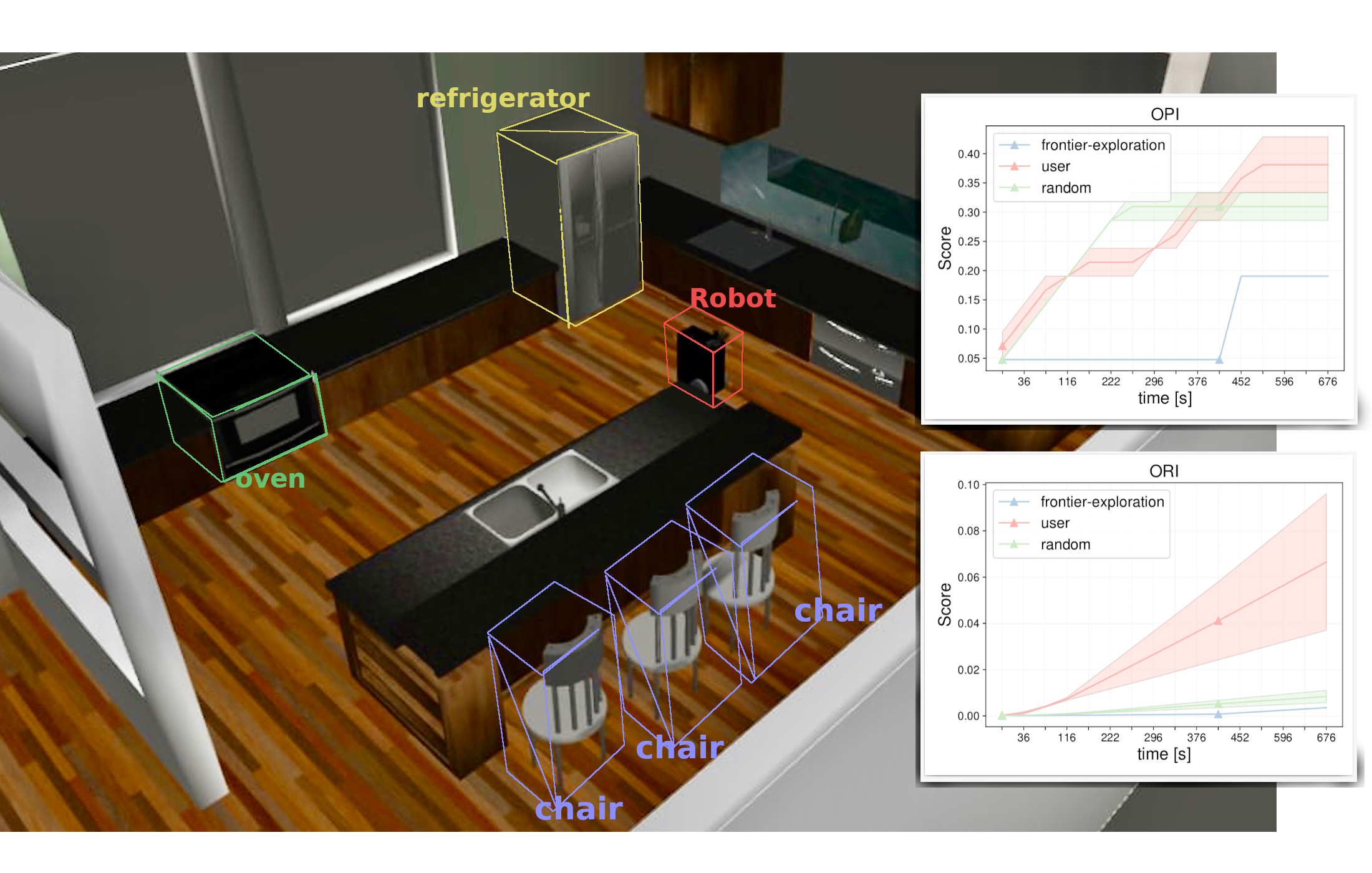}
    \caption{Robot performing semantic mapping within the \rosmeery{} benchmarking environment. The robot is highlighted in red while, mapped objects are labeled with their semantic category using different colours. On the right, benchmarking metrics are reported to monitor the semantic mapping session. It is important to highlight that the goal of \rosmeery{} is to evaluate how semantic knowledge is being gathered by the robotic agent and such metrics evaluate system allows for a quantitative measurement of the robot performance. In the session described by the figure, the robot employs different exploration policies and state-of-the-art object detector.} 
    \label{fig:intro}
\end{figure}

\figurename~\ref{fig:intro} shows the \rosmeery{} environment during a semantic mapping benchmarking session. In the scene, the robot is highlighted with a red label and it is equipped with a range sensor as well as a RGBD camera. Moreover, the framework comes with a state-of-the-art object detector node based on YOLO-v5\footnote{\url{https://github.com/ultralytics/yolov5}}. We report the output labels of such detector attached to the objects in the figure. Finally, in this example, the framework evaluates the robotic agent in accordance to two metrics: (1) the objects predicates recognized and (2) the accuracy in reconstructing the object geometry. On the right, we show two plots representing the profile of the two metrics chosen for this experiment. Within the \rosmeery{} framework, metrics are always a function of time, and report the performance of the robot from \textit{T=0}, the beginning of the benchmark, to the current time \textit{T=t}.

Summarizing, this contribution aims at providing a systematic evaluation of semantic maps based on~\cite{Capobianco2015} and implemented in \textit{Gazebo}, that measures the accuracy in the reconstruction of semantic knowledge collected by a robotic agent exploring an environment. In our first release, we provide the evaluation framework \rosmeery; a robot agent already equipped with a state-of-the-art object detector module; and a set of simulated environments as free-to-download packages which can be easily used to integrate new exploration strategies to perform benchmarking of semantic mapping approaches \footnote{\url{https://github.com/Lab-RoCoCo-Sapienza/S-Ave}}. To the best of our knowledge, this is the first available benchmark on robot semantic map exploration that also provides baseline methods.

The remainder of the paper is structured as follows. \sectionname~\ref{sec:related} contextualizes our work within the current state-of-the-art in semantic mapping focusing on its formalization and evaluation. Then, \sectionname~\ref{sec:framework} introduces the adopted semantic map formalization and describes in detail our benchmarking environment. \sectionname~\ref{sec:evaluation} describes the experimental evaluation setup and the simulated environment, presenting the obtained results. Finally, \sectionname~\ref{sec:conclusion} summarizes the key points of this paper, discusses its limitations and future research directions. 

\section{RELATED WORK}
\label{sec:related}
In this section, we contextualize our work within the state-of-the-art. First, we introduce few of the most known semantic mapping systems for mobile robots and then we survey proposed methodologies to evaluate them.

\subsection{Semantic Mapping}
The operative environment of a robot can be represented in multiple
ways, each of which can highlight different aspects and
features~\cite{Pronobis2017}. In this paper, we focus on a complete
representation of the world that spans different levels of abstraction
-- from raw sensor data to semantic human-level concepts. The problem
of representing the semantics of environments based on their spatial
location, geometry and appearance~\cite{Kostavelis2015} is usually
referred to as ``semantic mapping''. Semantic maps assign a certain
number of labels or properties to relevant features of the
environment~\cite{Goerke2009,Mozos2012}, and represent this knowledge in a usable way.

In literature, the wide heterogeneity of representations used for
semantic maps makes their comparison, evaluation and benchmarking
almost impossible. For example, Galindo et al.~\cite{Galindo2005} anchor sensor data, that describe rooms or objects in a spatial hierarchy, to the
corresponding symbol of a conceptual hierarchy in description logic
used by a robot. The authors validate their approach on a self-made
domestic-like environment and test the learned model by executing
navigation commands. Pangercic et al.~\cite{Pangercic2012}, instead, create semantic object maps by means of a knowledge base in description logic
associated to Prolog predicates (for inference). Such a knowledge base
contains classes and properties of objects, instances of semantic
classes and spatial information. The authors experiment their approach
on a PR2 robot which has to open a cabinet, and to detect handles based on a given semantic map. Riazuelo et al.~\cite{Riazuelo2015} make use of an ontology, for coding concepts and relations, and of a SLAM map for representing the scene geometry and object locations. Finally, Pronobis et al.~\cite{Pronobis2012}, represent a conceptual map as a probabilistic chain graph model and
evaluate their method by comparing the robot belief to be in a certain
location against the ground truth.

In order to evaluate semantic maps, Gunther et al.~\cite{Gunther2013} propose the usage of the rate of correctly classified objects. Handa et al.~\cite{Handa2014}, instead, propose a synthetic dataset, which can be extended with semantic knowledge and used as a groundtruth for comparing semantic mapping methods. Finally, Capobianco et al.~\cite{Capobianco2015} introduce a unified representation for semantic maps and pose the basis for their benchmarking system. In our work, as we discuss more in detail in the next section, we build upon~\cite{Capobianco2015} to provide a systematic evaluation of semantic maps and develop a benchmarking package.

\subsection{Semantic Maps Evaluation}
In order to evaluate the suitability and the effectiveness of a semantic map, several methodologies have been proposed in each of the paper contributions existing in the literature~\cite{Kostavelis2015}. And arguably, it is exactly the type of semantic maps evaluation that we are defying here by introducing the \rosmeery{} framework. For example in~\cite{Deeken2018}, the authors ground semantic maps in a spatial database and evaluate the correctness of the collected semantic knowledge against the groundtruth. However, such a particular implementation of the spatial database prevents any other semantic mapping algorithm, that does not feature a copy of such database, to be fairly compared. With a different algorithm, but with a similar results in terms of comparison with other approaches, the work in~\cite{Yue2020} introduces a hierarchical collaborative probabilistic semantic mapping algorithm that stores information by exploiting a voxel map, where each voxel also features a semantic label (e.g. floor, furniture). However, such a representation prevents any comparison with other characterizations of semantic entities and obfuscates metrics designed to evaluate accuracy in reconstructing objects within an environment. The authors of the paper do not provide a comparative evaluation with other 3D semantic object reconstruction algorithms. In fact, if we consider~\cite{Deeken2018}, it is already very difficult to compare the two approaches. Akin~\cite{Yue2020}, also the authors in~\cite{Crespo2020,Pronobis2017} adopt a hierarchical formalization of semantic knowledge. However, the layers of the hierarchies represent different levels of abstraction that intersect but not overlap -- including different semantic entities. For example in~\cite{Crespo2020}, the authors insert above the sensory layer an intermediate \textit{place layer} that keeps record of local nodes to support a topological representation of the environment, which is not present in~\cite{Yue2020}, that directly connects the sensory layer to a segmented representation of semantic objects. Also in these different setups, even though the goal is always to locate and label 3D objects, it becomes nearly impossible to assess the advantages and disadvantages of various semantic knowledge representations when brought into comparison. Additionally, it is important to remark that a proper comparison is possible only when there is a common understanding of the semantic knowledge to be reconstructed and represented. In fact, even though semantic knowledge features a staggering amount of different facets, it is important to adopt a general representation that can consider any type of semantic knowledge. To this end,  we recognize the proposal in~\cite{Capobianco2015} as a simple and general-enough semantic map representation that we can adopt in this work to perform benchmarking of different semantic mapping algorithms.\\
Summarizing, the aforementioned approaches provide an effective set of methodologies to ground abstract semantic labels into low-level robot sensory information. However, all these solutions are built with strong assumptions on the domain of application and too often describe an experimental evaluation that does compare to the state-of-the-art and evaluate the proposed solution in a way that is difficult to reproduce. Hence, to overcome these limitations, we contribute to the state-of-the-art with an evaluation benchmarking environment specifically designed for robot semantic mapping algorithms. Such a framework is designed to evaluate multiple aspects of such algorithms in accordance with custom metrics that are formalized in accordance with the comprehensive semantic map representation introduced in~\cite{Capobianco2015}.

\section{\rosmeery}
\label{sec:framework}
In this section, we introduce the semantic map representation that we adopt, as well as a detailed description of the \rosmeery{} framework and an initial set of metrics we decided to implement.

\subsection{Semantic Map Representation}
In order to build the \rosmeery{} framework, we adopt the formalization introduced in \cite{Capobianco2015}. Specifically, we consider a representation
composed by a tuple of three elements
\begin{equation}
    \set{SM} = \langle R, \set{M}, \set{P} \rangle,
    \label{eq:semantic_map}
\end{equation}
where:
\begin{itemize}
    \item $R$ is the global reference system in which all the elements of the semantic map are expressed;
    \item $\set{M}$ is a set of geometrical elements obtained as raw sensor data. They are expressed in the reference frame $R$ and describe spatial information in a mathematical form. For instance, if we consider a robot equipped with depth sensor, then $\set{M}$ can be instantiated to a point cloud representing the environment;
    \item $\set{P}$ is a set of predicates, among which \textit{is-a}(\texttt{X}, \texttt{Y}) and \textit{instance-of}(\texttt{X}, \texttt{Y}) are mandatory.
\end{itemize}

Here, the definition of a unique reference frame $R$ allows to associate the elements of $\set{M}$ with those in
$\set{P}$. Then, given two semantic maps $\set{SM}_1 = \langle R_{1},\set{M}_1,\set{P}_1 \rangle$ and $\set{SM}_{2} = \langle R_{2},\set{M}_{2},\set{P}_{2} \rangle$, an evaluation metric can be defined as
\begin{equation}
  \delta(\set{SM}_1,\set{SM}_{2}) = f(|\set{M}_1 \ominus \set{M}_{2}|, |\set{P}_1 \boxminus \set{P}_{2}|),
  \label{eq:metrics}
\end{equation}
where $R_{1}$ and $R_{2}$ must coincide (e.g., through a simple geometrical transformation). It is important to notice that since \equationname~\ref{eq:semantic_map} assumes a unique reference frame, it is particularly suitable for the analysis of different semantic maps. In fact, it allows to reformulate the problem of semantic maps comparison as the problem of anchoring their representation to a common reference frame. To tackle the issue, in this contribution, we provide an environment that ensures objects and their semantic characterization to belong to the same reference frame as the robotic agent maps them. Nevertheless, our ultimate goal is to expose a ready-to-use methodology to perform such a mapping and enable researches to embed their semantic map representation within the \rosmeery{} environment. We believe that research in this direction is key to enable proper benchmarking of inherently different semantic maps representations. Moreover, once they have been successfully transformed in a common reference frame, we can apply the same set of metrics (\equationname~\ref{eq:metrics}) in order to evaluate quantitatively the effectiveness of the semantic mapping algorithms. Such metrics can be applied at multiple level of abstraction, spanning from the low sensory level ($\ominus$) to a conceptual level ($\boxminus$) -- a more detailed description on how to implement such operators is provided in~\cite{Capobianco2015}. In this work, we introduce an initial and intuitive set of metrics, but it is worth remarking that the \rosmeery{} framework is designed to accept any set of metrics as input to the benchmarking session -- as long as such metrics are defined and implemented inline the determined reference frame.

\subsection{Architecture}
\label{sec:architecture}
\begin{figure}[!t]
    \centering
    \includegraphics[width=0.7\columnwidth]{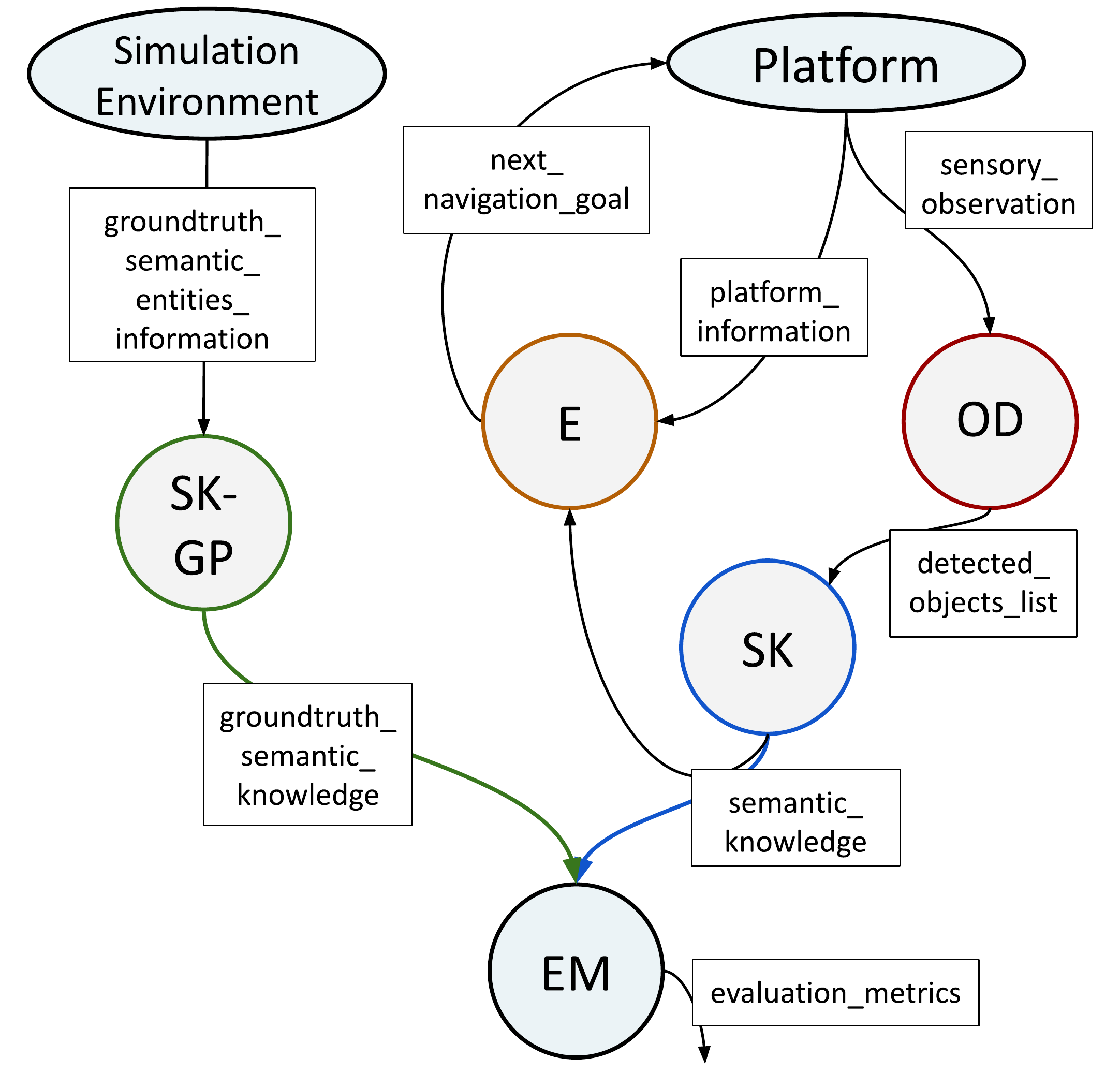}
    \caption{The \rosmeery{} architecture. Nodes are highlighted with different colors, while information exchanged among them is specified in boxes. Beginning at the top, both the Simulation Environment and the Platform node start two independent branches that join in the Evaluation metrics (EM) node (highlighted in black). Inputs of the EM node are two semantic knowledge instantiations of the same semantic map. One is complete and is stored in the Semantic Knowledge Groundtruth Provider (SK-GP, highlighted in green), that serves as a collector of groundtruth information directly from the simulation environment. The other instead, represents the accumulated knowledge collected by the robot during the benchmark, and stores it in the Semantic Knowledge (SK, highlighted in blue) node. Then, the semantic knowledge gathered during operation is detected and formalized with the aid of an object detector node (OD, highlighted in red) based on YOLO-v5. Finally, the architecture features an exploration node (E, highlighted in orange) that guides the robot in the environment and tries to optimize the semantic mapping.}
    \label{fig:rosmeery-arch}
\end{figure}

The \rosmeery{} framework is a benchmarking environment specifically designed to quantitatively evaluate semantic maps, and thus semantic mapping algorithms. To this end, we implement such a framework by relying upon the Gazebo simulation environment, that we exploit to reconstruct different areas of our university department. In fact, we demonstrate the flexibility of our framework that is agnostic to the actual scenario of application and can be deployed as long as a 3D simulation environment is provided, either in the case of real-life conditions~\cite{Deeken2018} or synthetically generated ones~\cite{Rosinol2020}. \figurename~\ref{fig:rosmeery-arch} illustrates the overall \rosmeery{} architecture by highlighting the active nodes and information exchanged among them. The figure shows two main \textit{independent} branches converging both to the evaluation metrics (EM) node that gathers both, the semantic knowledge accumulated by the robot since the beginning of the benchmark (blue arrow) and groundtruth knowledge extracted directly from the simulation environment (green arrow). As shown in the figure, the entire architecture can be schematically represented by a set of seven nodes, each of them designed with a specific task:

\begin{itemize}
\begin{figure}[!t]
    \centering
    \includegraphics[width=0.99\columnwidth]{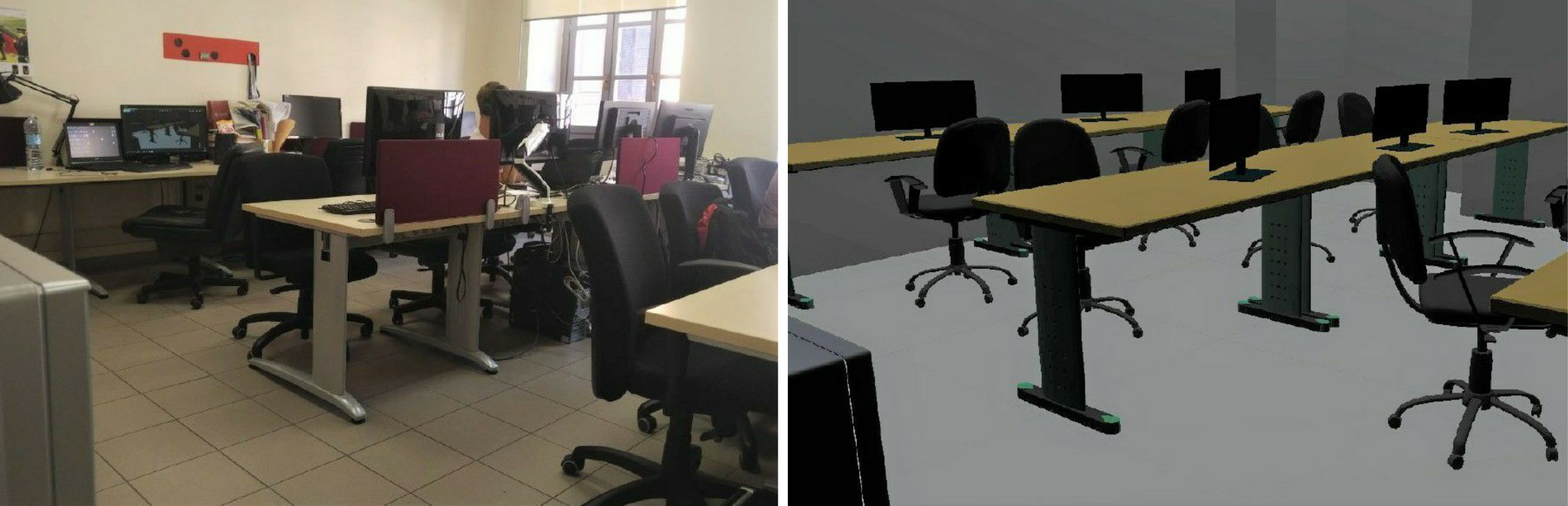}
    \caption{Comparison between a real world environment and its virtual replica.}
    \label{fig:sim-real-comparison}
\end{figure}

\begin{figure}[t!]
  \centering
  \subfigure[YOLO classification]{
    \includegraphics[width=0.7\columnwidth]{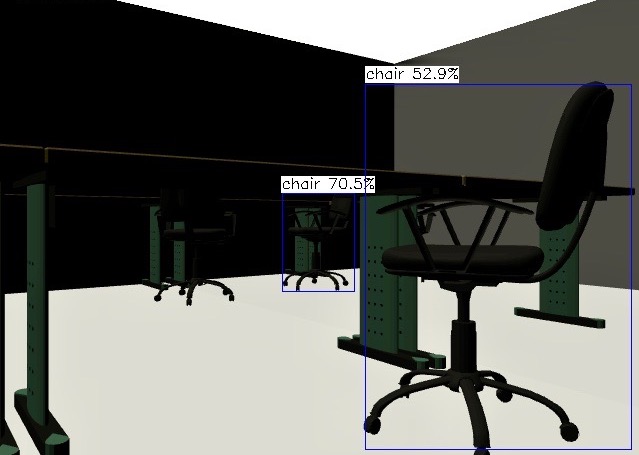}
    \label{fig:yolo-classification}
  }\\
  \subfigure[\rosmeery{} object detection message]{
    \includegraphics[width=0.8\columnwidth]{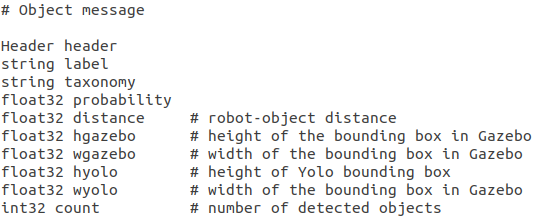}
    \label{fig:obj-message}
  }
  \caption{In the top corner, YOLO-v5 classification, showing objects detected from the robot position. In the bottom, fields of the object message used to populate the semantic knowledge database.}
  \label{fig:yolo-detection}
\end{figure}


    \item \textbf{Simulation Environment}. In order to enable proper benchmarking, it is fundamental to deploy the target algorithms in a controlled environment -- before moving to a real world scenario. In robotics, there is a remarkable amount of simulation environments that support research in several areas and provide great benefits. Accordingly, we present a benchmarking framework based on the Gazebo simulator where we can design and take control of the challenge that the robot is going to face when deployed in the \rosmeery{} framework. The environment is loaded each time a benchmark session is started and automatically places a set of pre-configured objects to populate the chosen scenario. \figurename~\ref{fig:sim-real-comparison} shows an example of an environment that we include in the repository. Importantly, the simulated worlds that we provide are all reconstruction of real-life location within our department. On the left of the figure, we can see a picture of a real room, while on the right, the same area reproduced virtually. Of course, it is not a mandatory requirement, and virtual scenes can also be synthetically generated. The effort we put in reproducing real-life environments rewards us with the possibility to also execute the benchmarking session with a real robot navigating and building a semantic map of our department.
    \item \textbf{Semantic Knowledge Groundtruth Provider (SK-GP)}. Intuitively, this node (highlighted in green) is an important asset of the overall architecture, and serves as a collector of the groundtruth semantic map used as target reference. SK-GP has a first interaction with the simulated world and keeps a list of semantic entities attributes existing in the environment, such as location, size, and semantic label. Input to SK-GP is a list of descriptors of the objects occurring in a particular run of the benchmark that are mandatory in order to compute the desired metrics. In fact, the main task of such a node is to ensure that all the information needed to evaluate the robotic agent is present and exposed from the simulator. Generally, the node stores semantic knowledge in accordance with the semantic map representation introduced in~\equationname~\ref{eq:semantic_map} and outputs a list of information where each element relates to a particular object and provides both geometrical and appearance ($\set{M}$) information, together with conceptual objects predicates and attributes ($\set{P}$)~\cite{Capobianco2015}.
    \item \textbf{Platform}. The other branch of the architecture originates from the platform node. Such a node represents the robotic platform, equipped with a set of desired sensors, and can be operated with velocities commands to navigate the environment. In this release, we provide a simple robot equipped with a laser range sensor and a RGBD camera, however, Gazebo offers a variety of sensors that can be included. The platform provides low-level sensory observation that are used to both, extract semantic knowledge from the environment and command the robot movements. In fact, the node accepts target poses as input, and triggers the navigation stack to reach them. Such target poses are generated by the exploration node that can be instantiated to any map exploration algorithm. Also in this case, we virtually reproduce a custom-made real robot but, any platform can be used, as long as it obeys physical embodiment constraints in a real-world setting. 
    \item \textbf{Exploration (E)}. The exploration node (highlighted in orange) is the first node belonging to the \rosmeery{} core, which is a set of three nodes implementing the sense-plan-act paradigm~\cite{DeSilva2008}. The policy that the robot exploits to discover semantic knowledge can actively affect the overall performance both, in terms of accuracy and time to explore an area. Importantly, such a node must serve as an interface that assumes as input the robot status to output the next best-pose to reach in order to increment the semantic understanding on the environment. In this implementation, we provide three different instantiations of the exploration node: a frontier based exploration policy, where the robot always selects the biggest frontier of the map already explored; a random walker, where the robot selects the next robot pose randomly in the environment; and a human-driven exploration policy, where the robot is commanded by a user operator. In the latter setting, users have been instructed to maximize a set of metrics describing the semantic mapping task, however, in order to provide a fair comparison with other approaches, users could only observe the environment through a top-down 2D map and the RGBD camera mounted on the robotic platform. As we report in the experimental evaluation section, each experiment has been repeated several times with different users to guarantee statistical significance.
    \item \textbf{Object Detector (OD)}. The \rosmeery{} framework comes with a state-of-the-art object detector node called YOLO-v5 (red in the figure). However, such a node can be disabled at the beginning of each benchmarking session as we understand that many semantic mapping systems already feature an object detection module. Specifically, the OD we provide is based on YOLO-v5, which comes with 4 different architectures depending on their model sizes (S, M, L and X). YOLO-v5 implements the CSP Bottleneck~\cite{wang_liao_wu_chen_hsieh_yeh_2020} to formulate image features as backbone. The CSP models are based on DenseNet. PA-NET~\cite{liu_path_2018} was implemented as neck for feature aggregation. YOLO-v5 uses data augmentation and auto-learning for the bounding box anchors. The object detector node is deployed by adapting ROS NCNN package~\cite{tencent}, which is a high-performance neural network inference framework developed by Tencent, optimized for mobile platforms and ROS. Such a package provides the feasibility to change the Object Detection algorithm. We used YOLO-v5(L) for the experimentation, as this model achieved AP-val of 48.4, AP-test of 48.4 and AP-50 of 66.9. The network was trained on MS COCO dataset consisting of 80 classes for object detection. 
    Moreover, when OD is enabled, at each iteration it provides a list of detected objects with their bounding box, estimated location and semantic label associated to a confidence score as shown in \figurename~\ref{fig:yolo-classification}. 
    \item \textbf{Spatial Knowledge (SK)}. The spatial knowledge node (SK, blue in the figure) is in charge of receiving the processed sensory observations and to store them in the semantic map, built incrementally. In accordance with~\equationname~\ref{eq:semantic_map}, in this release, we propose a semantic mapping system that stores the set of geometrical elements $\set{M}$ as a segmented point cloud, by relying on the 3D SLAM system introduced in~\cite{Colosi2020}. A taxonomy, instead, is used to keep track of the set of predicates $\set{P}$ discovered each time a new object is added to the map. Such a predicates indicate the categories of elements and their relations (e.g. chairs are furniture, laptops are accessories, chairs are likely to be found nearby a table)~\cite{Gemignani2016}.  The association of each object detected by the robot with its category is obtained through a taxonomy, handcrafted in this release. The entries in the taxonomy are used to generate an \textit{Object message} (see~\figurename~\ref{fig:obj-message}) of each detected model that will be finally used to populated the robot semantic knowledge database.
    \item \textbf{Evaluation Metrics (EM)}. The evaluation metrics node (EM, black in the figure) is one of the most important building block within the \rosmeery{} framework. The node is initialized with a set of desired metrics functions that characterize the current benchmarking session. Metrics can be defined to evaluate both geometrical aspects and accuracy in the semantic map reconstruction ($\set{M}$) and at the predicate level ($\set{P}$). The latter case is usually neglected by most papers that, too often, only report accuracy of the mapped objects. However, we argue that a semantic map is complete only if it is associated to a relation graph that specifies predicates of the semantic knowledge discovered. Thus, the robotic agent should be evaluated also, and foremost, by considering the amount of predicates it can assess from its operation. For example, that a table is not only a table but also a piece of furniture. We believe that such category of information is fundamental for a service robot and, in order to optimize to a particular environment, such predicates must be learned during the operation. Hence, as we describe in the next section, we include in the initial set, a metric that explicitly evaluates the number of predicates the robot discovers in the environment. It is important to remark that the set of active metrics can be updated and incremented before each benchmarking session and can also be customized to specifically evaluate an aspect of the semantic mapping algorithm. Metrics measures are computed in accordance with~\equationname~\ref{eq:metrics}, that allows comparison with different semantic mapping algorithms and with different runs of the same.
\end{itemize}

Finally, we release \rosmeery{} with a pre-defined representation of the semantic maps that are compliant with~\equationname~\ref{eq:semantic_map} and it is a general enough description that can be used as an interface to access any representation adopted in the literature. In future releases, we will add a plugin to showcase how to embed specific semantic maps and to output semantic knowledge that is compliant with~\equationname~\ref{eq:metrics}.

\subsection{Metrics}
\label{sec:metrics}
We deliver our framework with two pre-defined metrics. This initial set of metrics is designed to showcase how both geometrical accuracy and discovered semantic predicates can be quantitatively measured -- referred to as \textit{object reconstruction index} (ORI) and \textit{object predicates index} (OPI) respectively. Such metrics are an instantiation of \equationname~\ref{eq:metrics}, are in $[0, 1]$ and are computed against the groundtruth of the environments, which contains the true number of objects and predicates; and objects sizes computed on the 3D point cloud.

\paragraph{ORI.} The object reconstruction index is designed to evaluate the reconstruction accuracy in representing objects perceived during operation. As mentioned previously, we keep a description of the geometrical aspects of the environment as a segmented point cloud. Thus, we can compare the number of points assigned to a particular object against the groundtruth and, for each detected object, we report an average of the amount of surface that has been observed and added to the point cloud. We compute the ORI index in two ways: the former is a flat comparison of the segmented points clouds averaged over all the objects existing in the scene. The latter, considers the noisy object detection output and weights the reconstructed point cloud by exploiting the object label confidence score (outputted by YOLO-v5). Intuitively, in order to achieve high metric scores, such a metric forces the semantic mapping algorithm to be accurate and confident in detecting objects. Within \rosmeery{} we configure these two implementations as two different metrics that we refer to as ORI~\equationname~\ref{eq:ori}, and confidence-based ORI (cORI)~\equationname~\ref{eq:cori}. ORI is computed as follows
\begin{equation}
  ORI = f(|ps(\cdot) \ominus ps_G(\cdot)|, \emptyset) = 1 - min\left(1,\frac{1}{N}\sum_{n=1}^N{\frac{\abs{ps_G(n) - ps(n)}}{ps_G(n)}}\right)
  \label{eq:ori}
\end{equation}
where $f(\cdot)$ is the evaluation function introduced in~\equationname~\ref{eq:metrics}, $n$ denotes the n-\textit{th} object in the scene, $N$ is the total number of objects, $ps_G(\cdot)$ is the groundtruth number of points in the point cloud surface representing $n$, and $ps(\cdot)$ is a function that returns the number of perceived surface points of $n$ accumulated during the benchmark. Similarly, cORI is computed as a weighted average sum that also considers object confidence scores as follows
\begin{equation}
  cORI = f(|c(\cdot)ps(\cdot) \ominus ps_G(\cdot)|, \emptyset)| = 1 - min\left(1,\frac{1}{N}\sum_{n=1}^N{\frac{\abs{ps_G(n) - (c(n)\cdot ps(n))}}{ps_G(n)}}\right)
  \label{eq:cori}
\end{equation}
where $f(\cdot)$ is the evaluation function introduced in~\equationname~\ref{eq:metrics}. $n$, $N$, $ps(\cdot)$ and $ps_G(\cdot)$ are as previously described in~\equationname~\ref{eq:ori} and $c(\cdot)$ is the confidence score outputted by the object detector OD and associated to $n$. It is worth remarking that we introduce such a metric to highlight the importance of taking into consideration the uncertainty in the robot detection and to mimic noisy sensory data of real-world setting. In fact we expect such a metric to always underestimate the standard ORI computation -- that always assumes a correct object classification. 

\begin{figure}[!t]
    \centering
    \includegraphics[width=0.99\columnwidth]{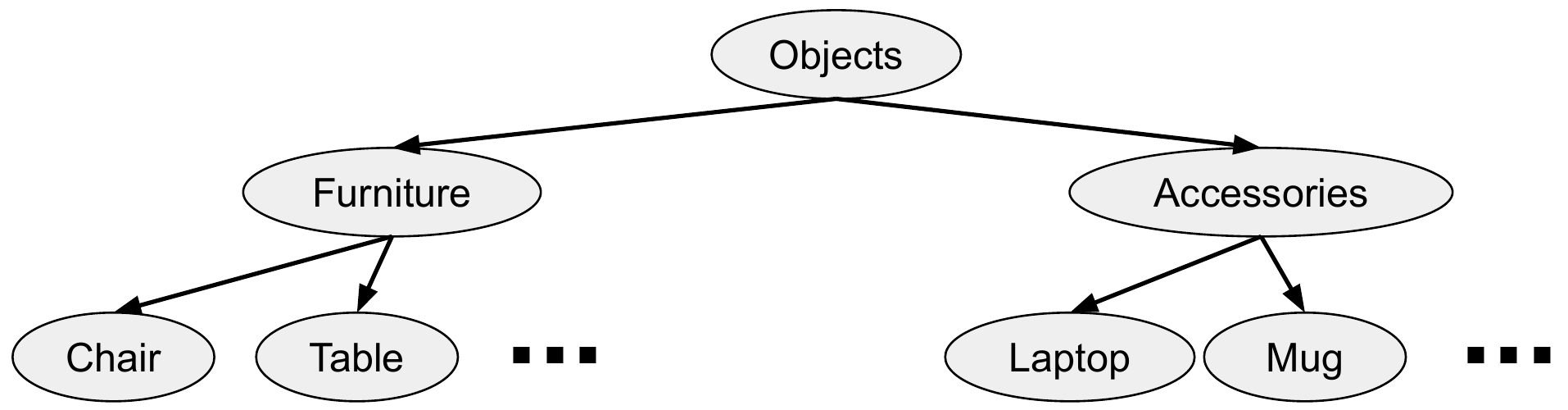}
    \caption{Object taxonomy.}
    \label{fig:obj_taxonomy}
\end{figure}
\paragraph{OPI.} The object predicates index computes the difference between the number of object predicates inferred in each trial $p$, and the total number of predicates $P$ that can be inferred by considering a target groundtruth taxonomy of the environment. Such a metric is computed in accordance with ~\equationname~\ref{eq:opi}:
\begin{equation}
  OPI = f(\emptyset, |p(\cdot) \boxminus P_G(\cdot)|) = 1 - min\left(1,\frac{1}{N}\sum_{n=1}^N {\frac{\abs{P_G(n) - p(n)}}{P_G(n)}}\right)
  \label{eq:opi}
\end{equation}
where $f(\cdot)$ is the evaluation function introduced in~\equationname~\ref{eq:metrics}, $n$ is the n-\textit{th} object in the scene, $P_G(\cdot)$ is the groundtruth number of predicates associated to $n$, and $p(\cdot)$ is the number of predicates declared by the agent and stored in the set $\set{P}$ (see~\equationname~\ref{eq:semantic_map}). As shown in the example in~\figurename~\ref{fig:obj_taxonomy}, in this contribution we consider a basic taxonomy that tracks object categories and macro-categories. However, the definition of the metric can include an arbitrary number of predicates, as long as, those are also implemented in the SK-GP node. Moreover, in the case in which two or more objects are detected in the same frame, and are possibly occluding each other, the ORI and OPI metrics are update by considering all the detected models. In fact, given the non-ideal object detector we are employing, we can configure a minimum threshold for the confidence value of each detection, and thus, we can include in the computation of the metrics only high-confidence objects which suggestively discards ambiguous and too occluded models in the comparison with the groundtruth. Also in the case of multiple objects clustered together and detected as one, the two metrics proposed will account for such a misclassfication as the OPI would result in a lower number of predicates discovered and ORI would decrease the score since there is a mismatch between the computed object surfaces and the groundtruth. However, it is worth highlighting that the metrics proposed here are used to showcase the potential of our framework and more sophisticated evaluation criteria can be integrated.

Finally, the evaluation $f(\cdot)$ introduced in~\equationname~\ref{eq:metrics} can also be represented by considering all the introduced metrics in order to evaluate the benchmarking session holistically. Moreover, in this proposal the OPI index is implemented as a difference between the discovered predicates and the groundtruth. However, a more sophisticated evaluation of semantic knowledge can be carried out by considering the knowledge graph that the agent can build. In fact, the $\boxminus$ can be also implemented as the ``semantic similarity'' index introduced by the authors of~\cite{Maxat2020}.

\section{EVALUATION AND BENCHMARKING}
\label{sec:evaluation}
\begin{figure}[t!]
  \centering
  \subfigure[env 0]{
    \includegraphics[width=0.45\columnwidth]{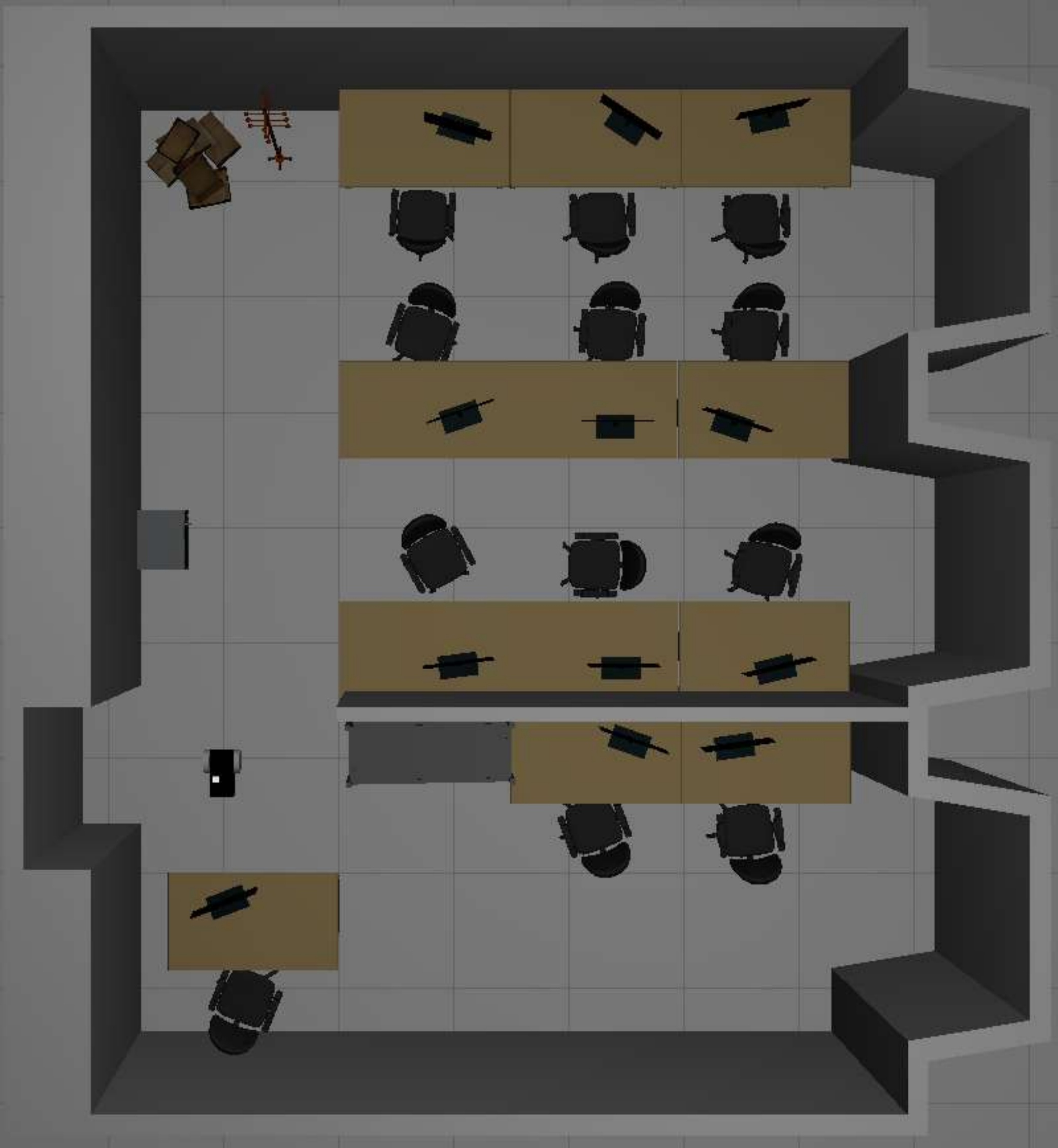}
    \label{fig:env-3}
  }
  \subfigure[env 1]{
    \includegraphics[width=0.45\columnwidth]{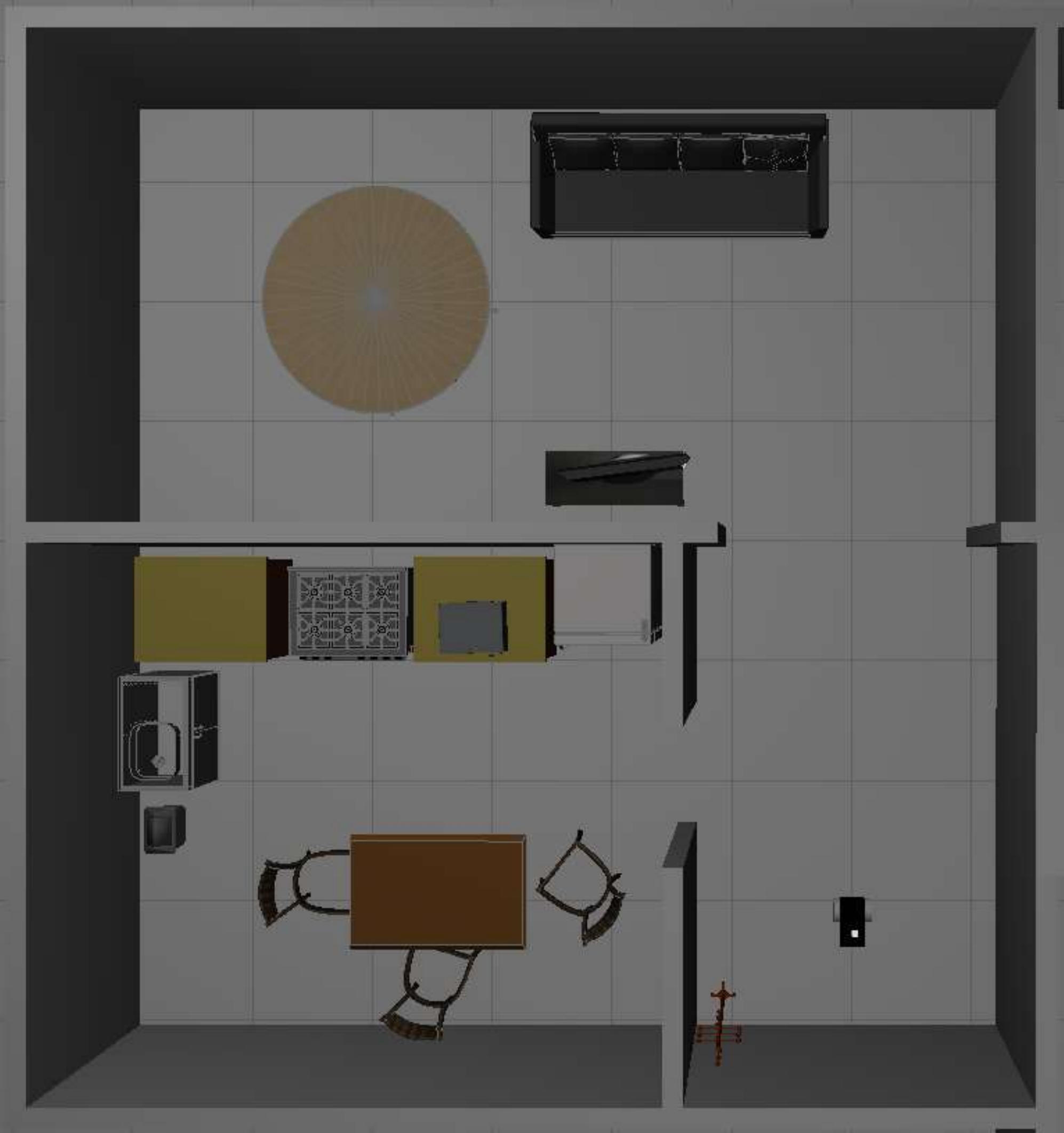}
    \label{fig:env-4}
  }
  \caption{Indoor environments used to evaluate the different approaches. Environments differ in terms of structure, topology and objects within.}
  \label{fig:environments}
\end{figure}

\begin{figure}[t!]
  \centering
  \subfigure[Kitchen - ORI]{
    \includegraphics[width=0.47\columnwidth]{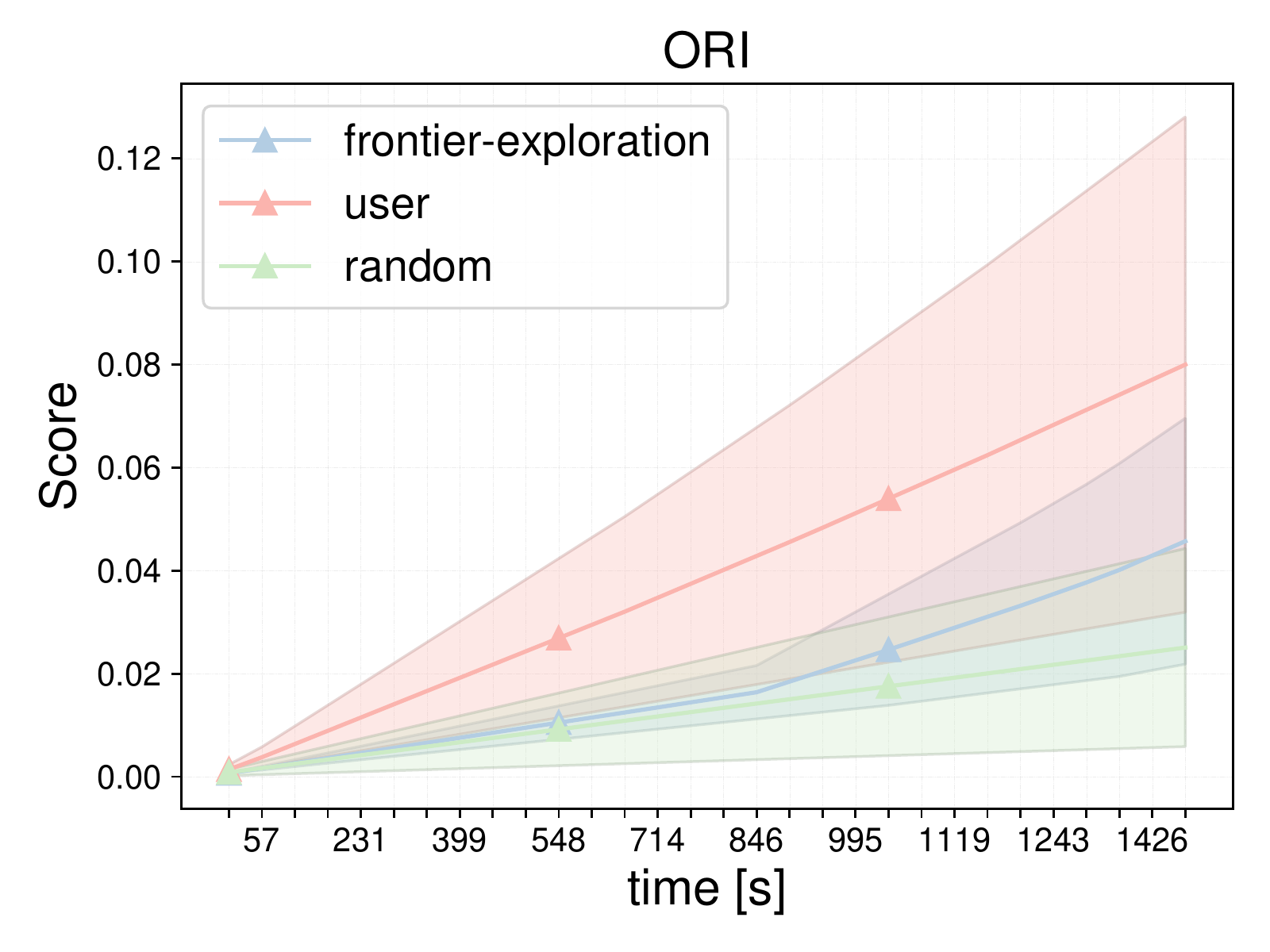}
    \label{fig:benchmarking-plots-kitchen-4-ori}
  }
  \subfigure[Kitchen - cORI]{
    \includegraphics[width=0.47\columnwidth]{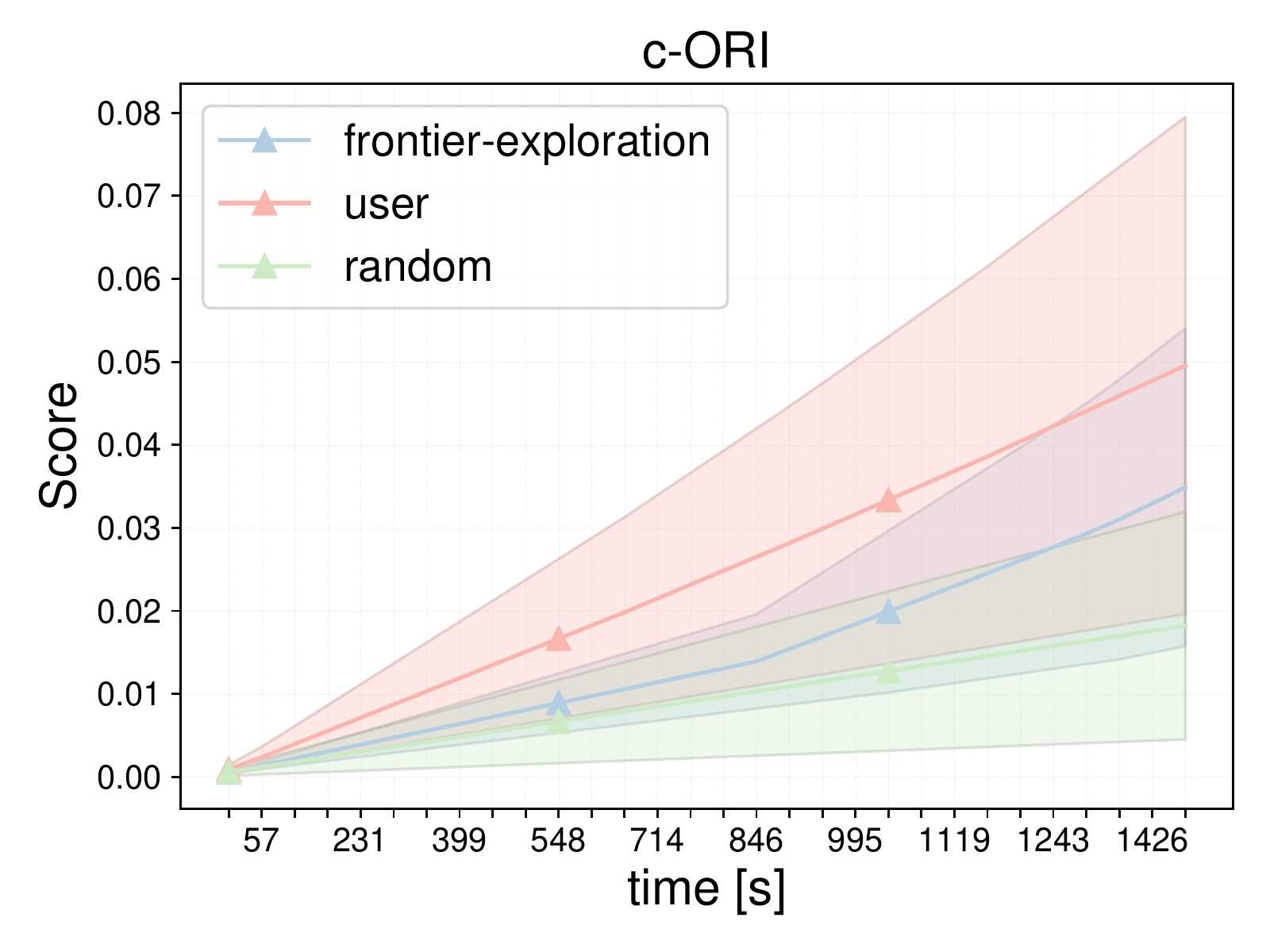}
    \label{fig:benchmarking-plots-kitchen-4-cori}
  }\\
  \subfigure[Kitchen - OPI]{
    \includegraphics[width=0.47\columnwidth]{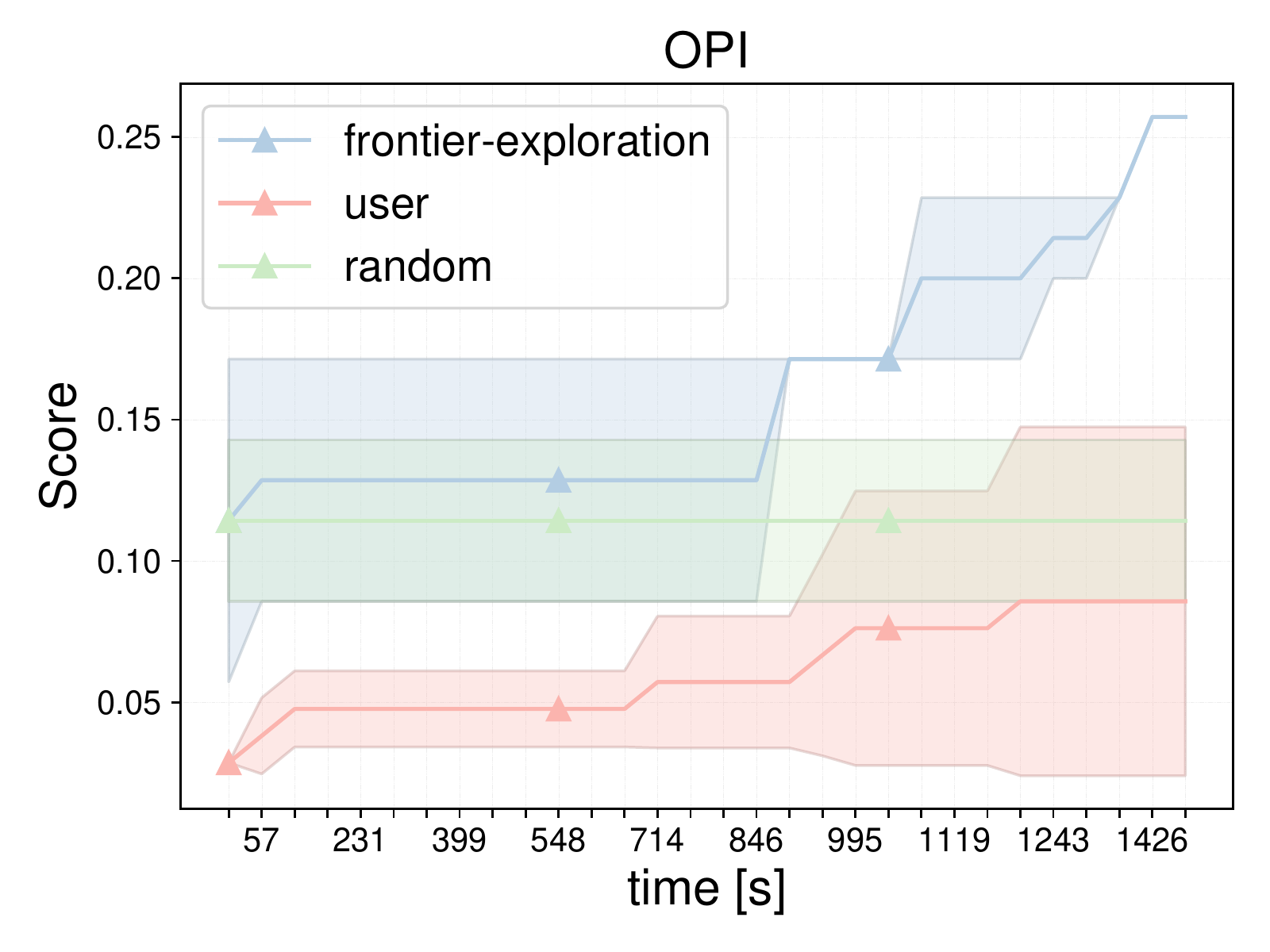}
    \label{fig:benchmarking-plots-kitchen-4-opi}
  }
  \subfigure[Laboratory - ORI]{
    \includegraphics[width=0.47\columnwidth]{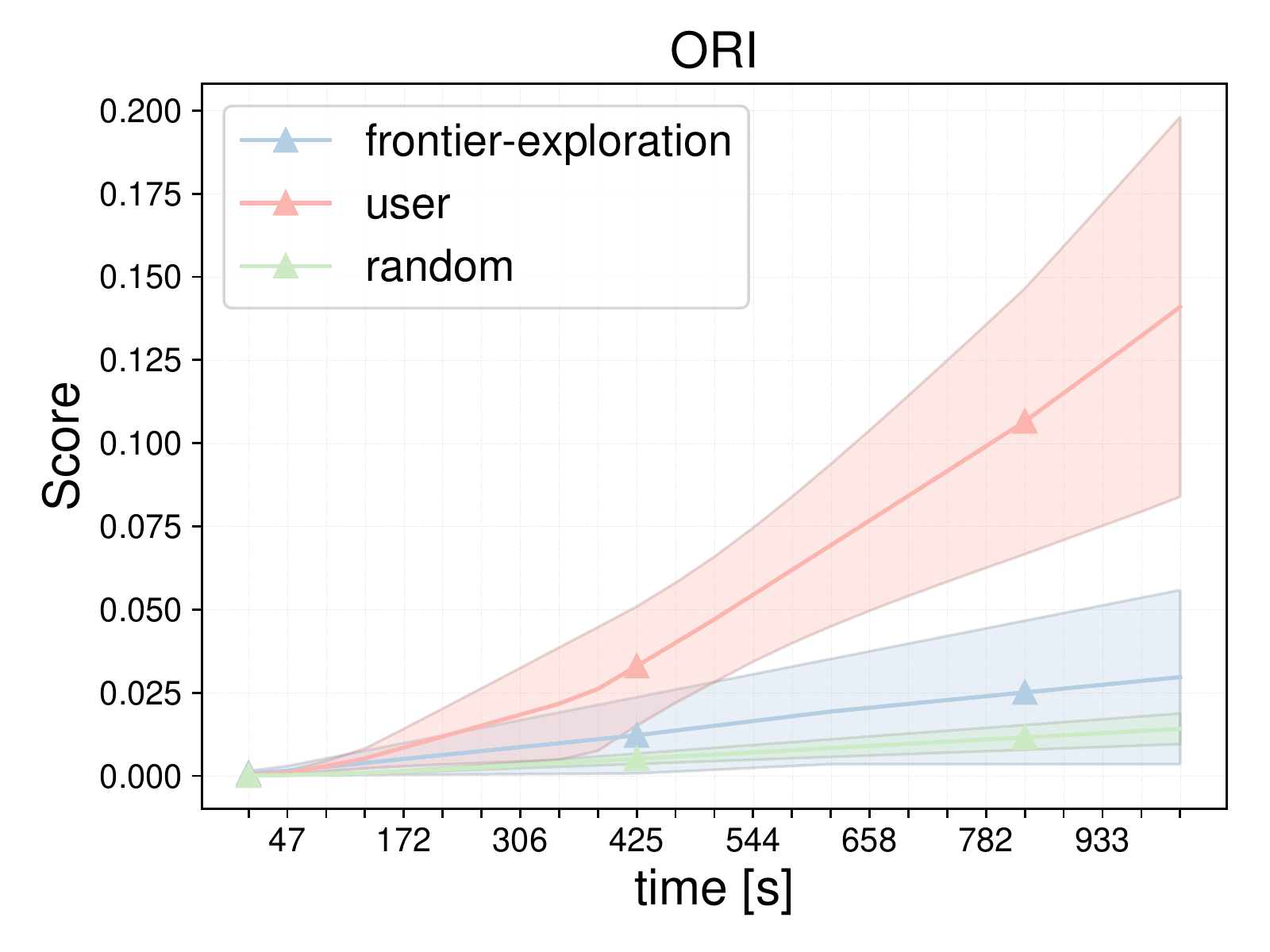}
    \label{fig:benchmarking-plots-laboratory-4-ori}
  }\\
  \subfigure[Laboratory - cORI]{
    \includegraphics[width=0.47\columnwidth]{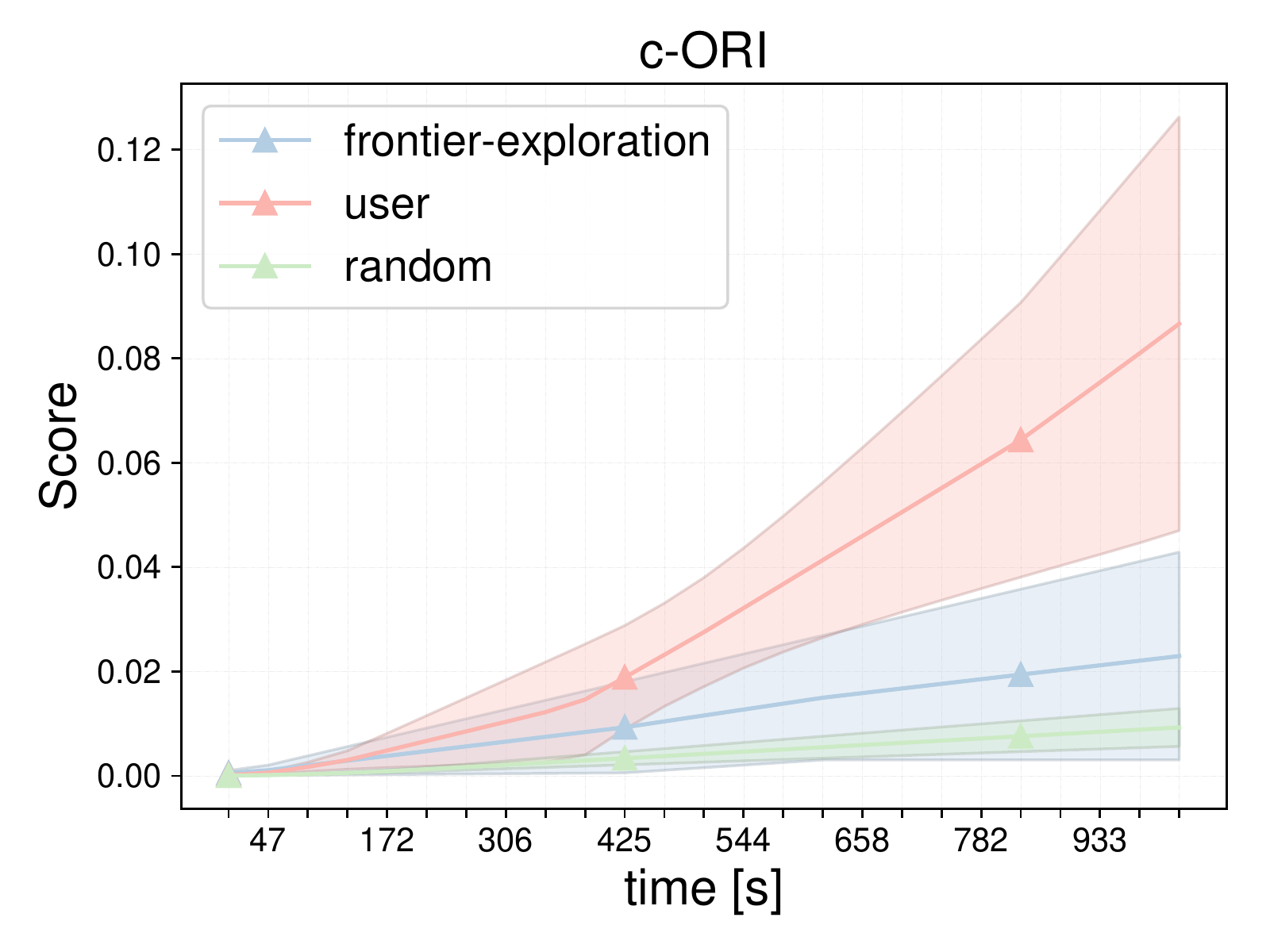}
    \label{fig:benchmarking-plots-laboratory-cori}
  }
  \subfigure[Laboratory - OPI]{
    \includegraphics[width=0.47\columnwidth]{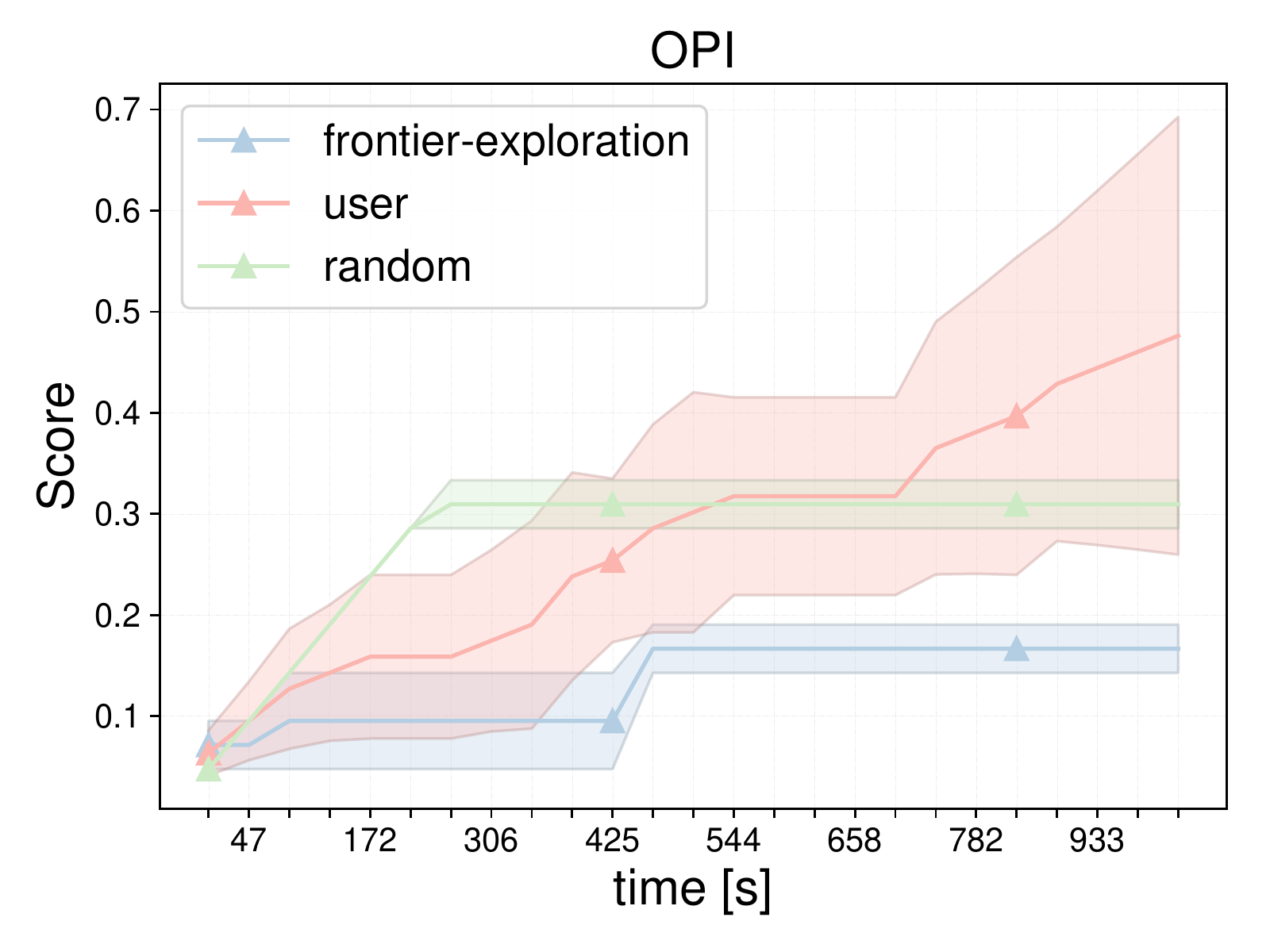}
    \label{fig:benchmarking-plots-laboratory-opi}
  }\\
  \caption{Benchmarking metric scores in the Kitchen and Laboratory environments. The exploration have been conducted adopting different policies: user-driven exploration (\textit{user}); frontier-based exploration (\textit{frontier-based exploration}) and random-walk exploration (\textit{random}). On the y-axis the metric score, while on the x-axis the time of the semantic mapping session in seconds. For each algorithm, the solid line represents the average score, while the shaded area is the standard deviation computed over 5 runs of the experiment.}
  \label{fig:benchmarking-plots-1}
\end{figure}

\begin{figure}[t!]
  \centering
  \subfigure[Small office - ORI]{
    \includegraphics[width=0.47\columnwidth]{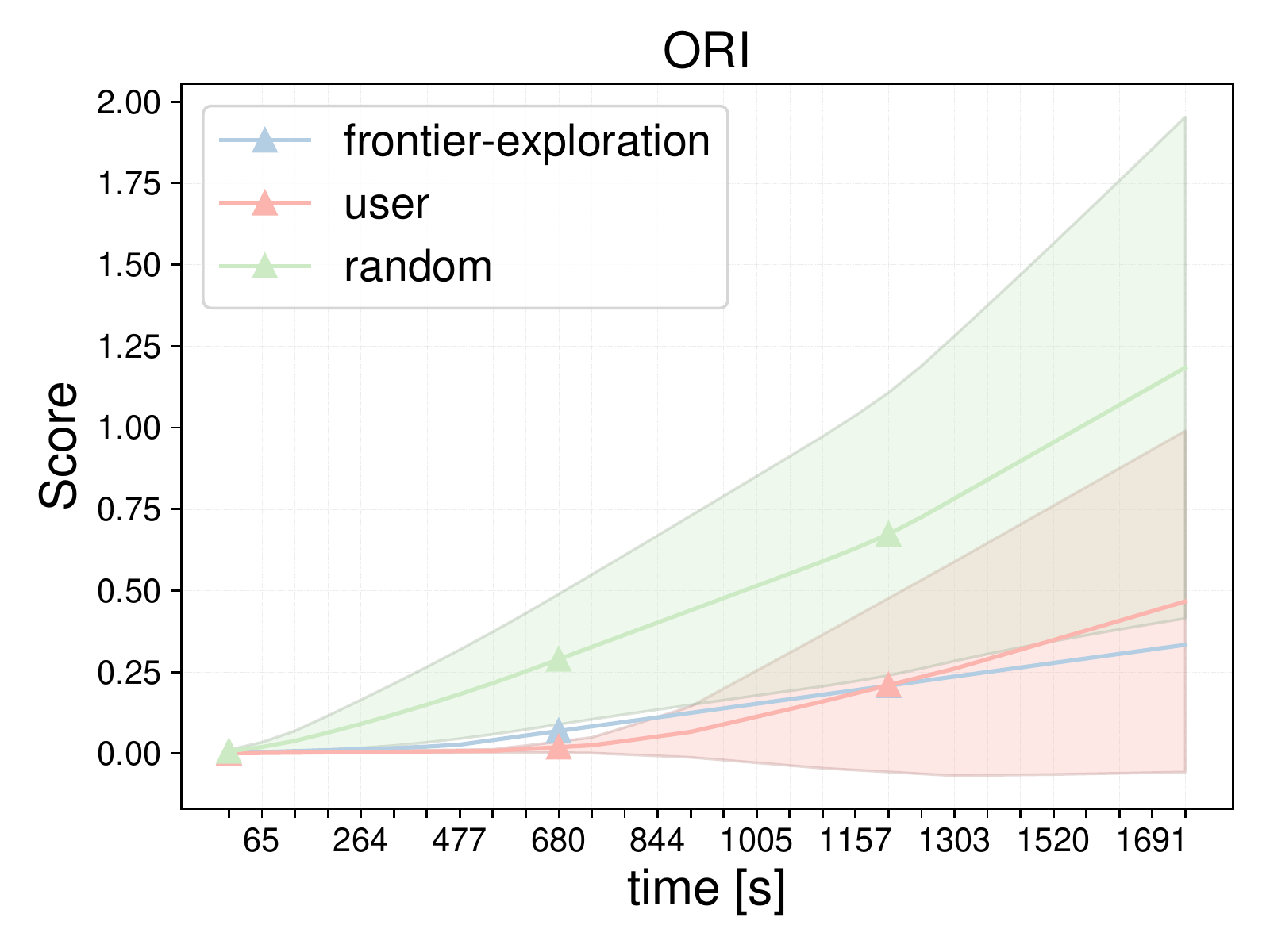}
    \label{fig:benchmarking-plots-small-office-ori}
  }
  \subfigure[Small office - cORI]{
    \includegraphics[width=0.47\columnwidth]{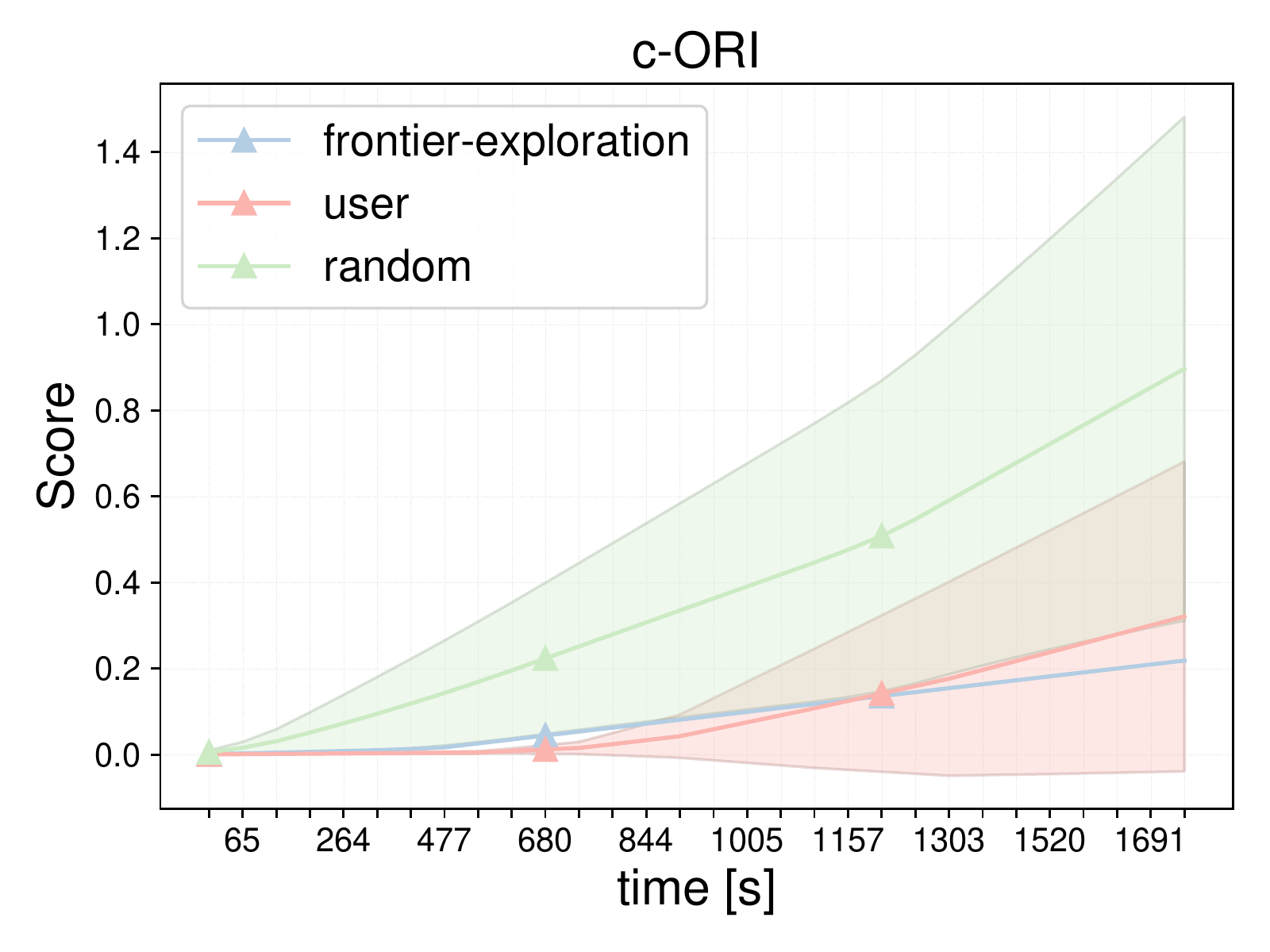}
    \label{fig:benchmarking-plots-small-office-cori}
  }\\
  \subfigure[Small office - OPI]{
    \includegraphics[width=0.47\columnwidth]{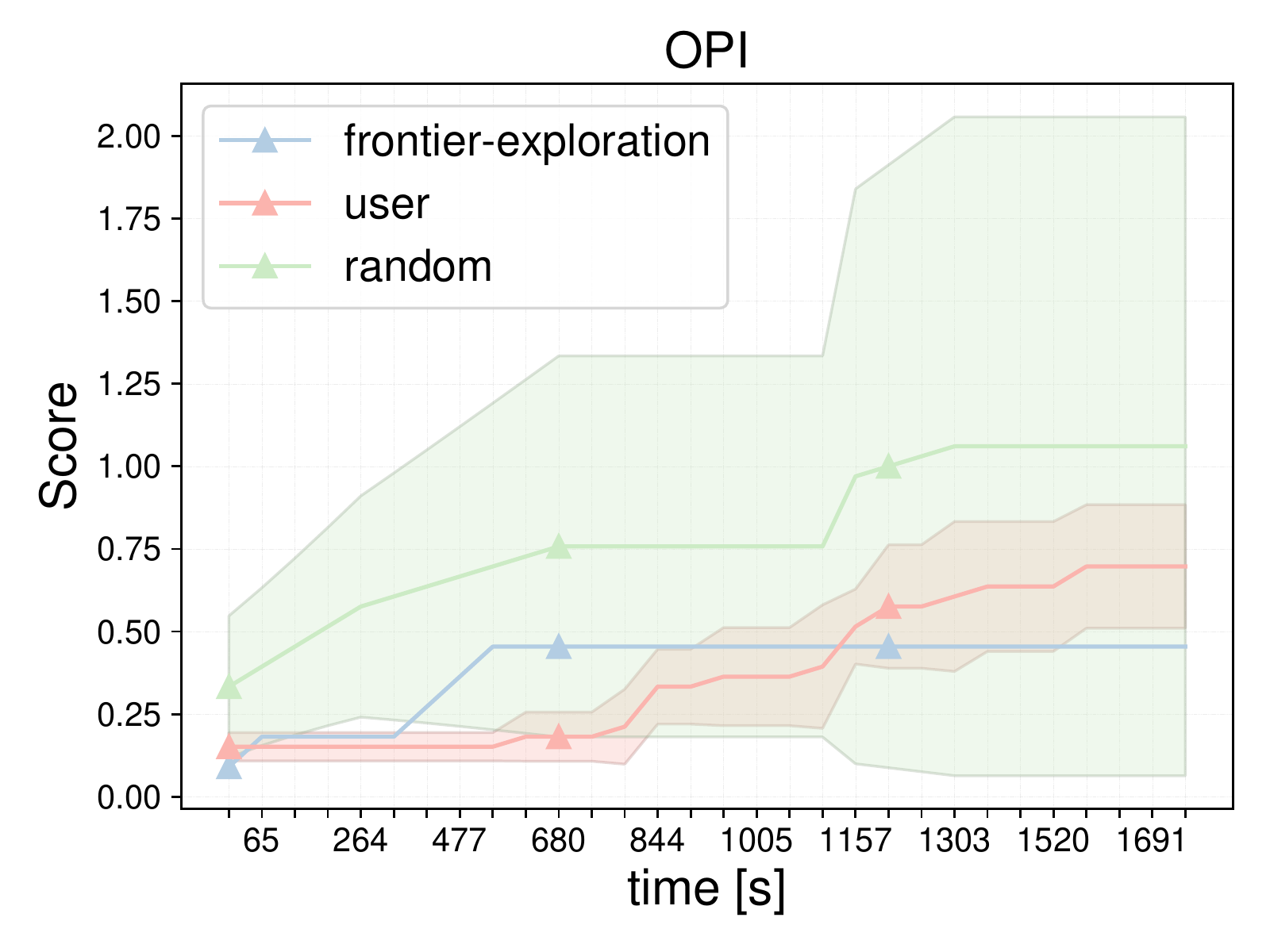}
    \label{fig:benchmarking-plots-small-office-opi}
  }
  \subfigure[Large office - ORI]{
    \includegraphics[width=0.47\columnwidth]{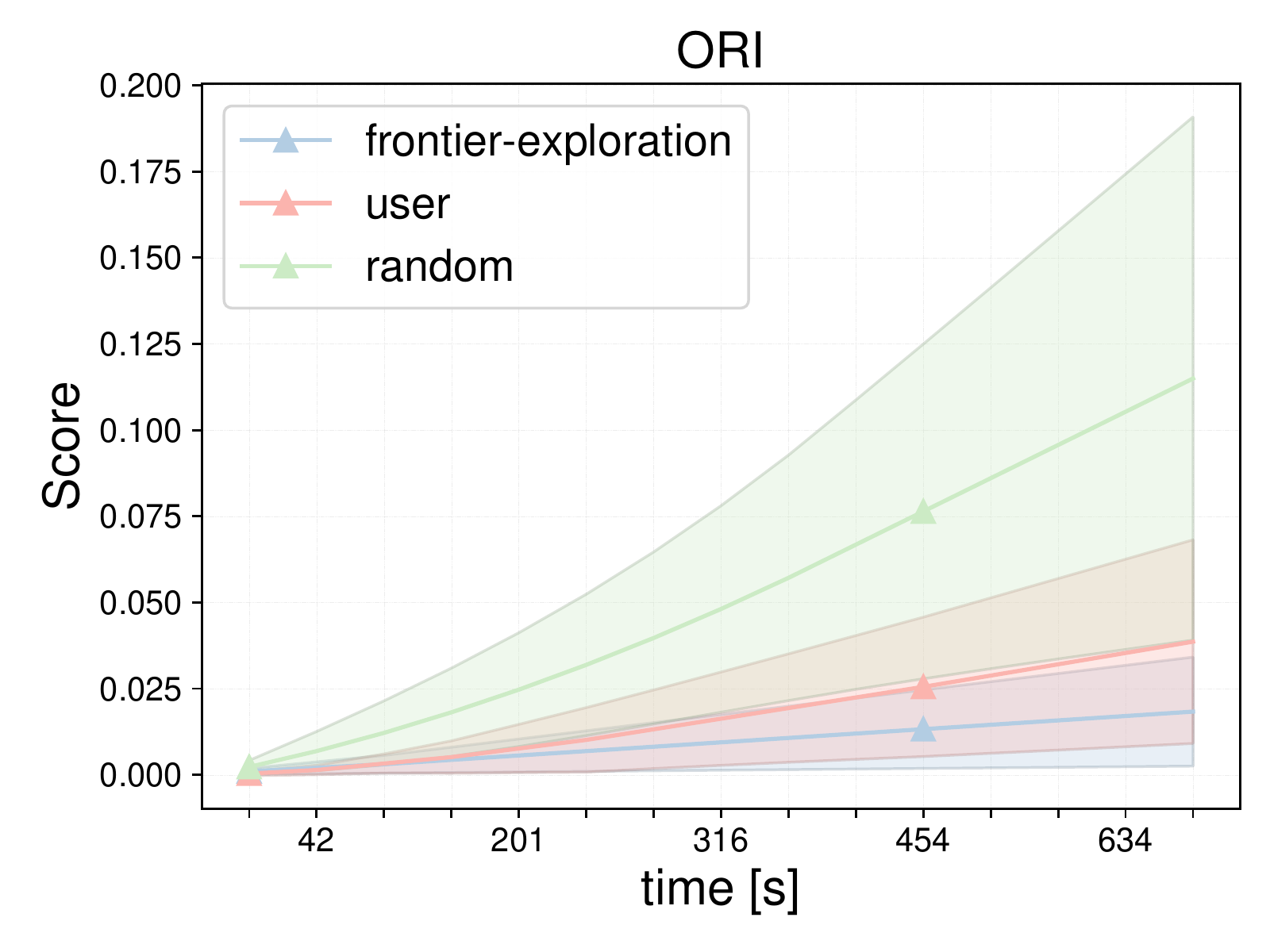}
    \label{fig:benchmarking-plots-large-office-ori}
  }\\
  \subfigure[Large office - cORI]{
    \includegraphics[width=0.47\columnwidth]{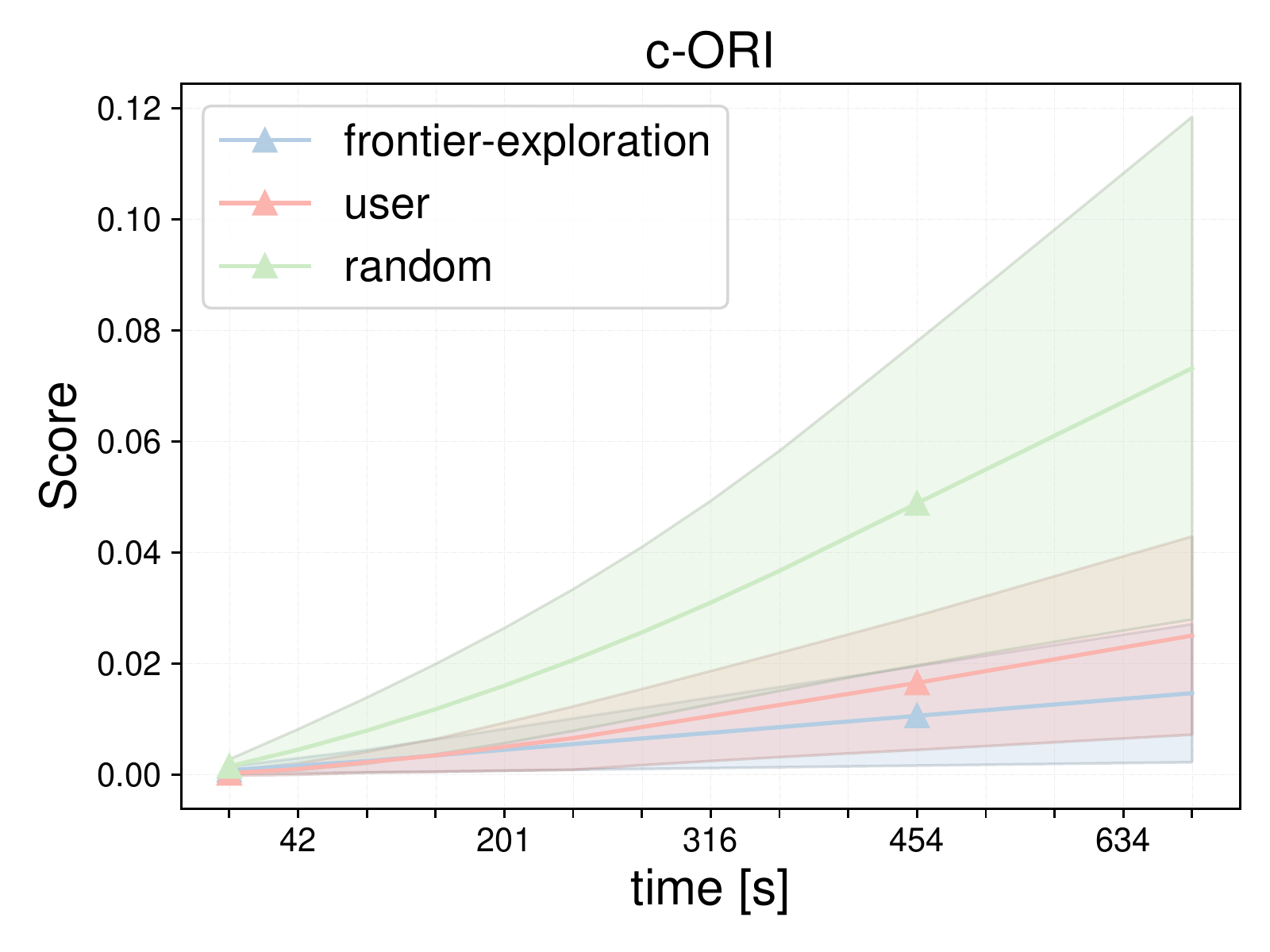}
    \label{fig:benchmarking-plots-large-office-cori}
  }
  \subfigure[Large office - OPI]{
    \includegraphics[width=0.47\columnwidth]{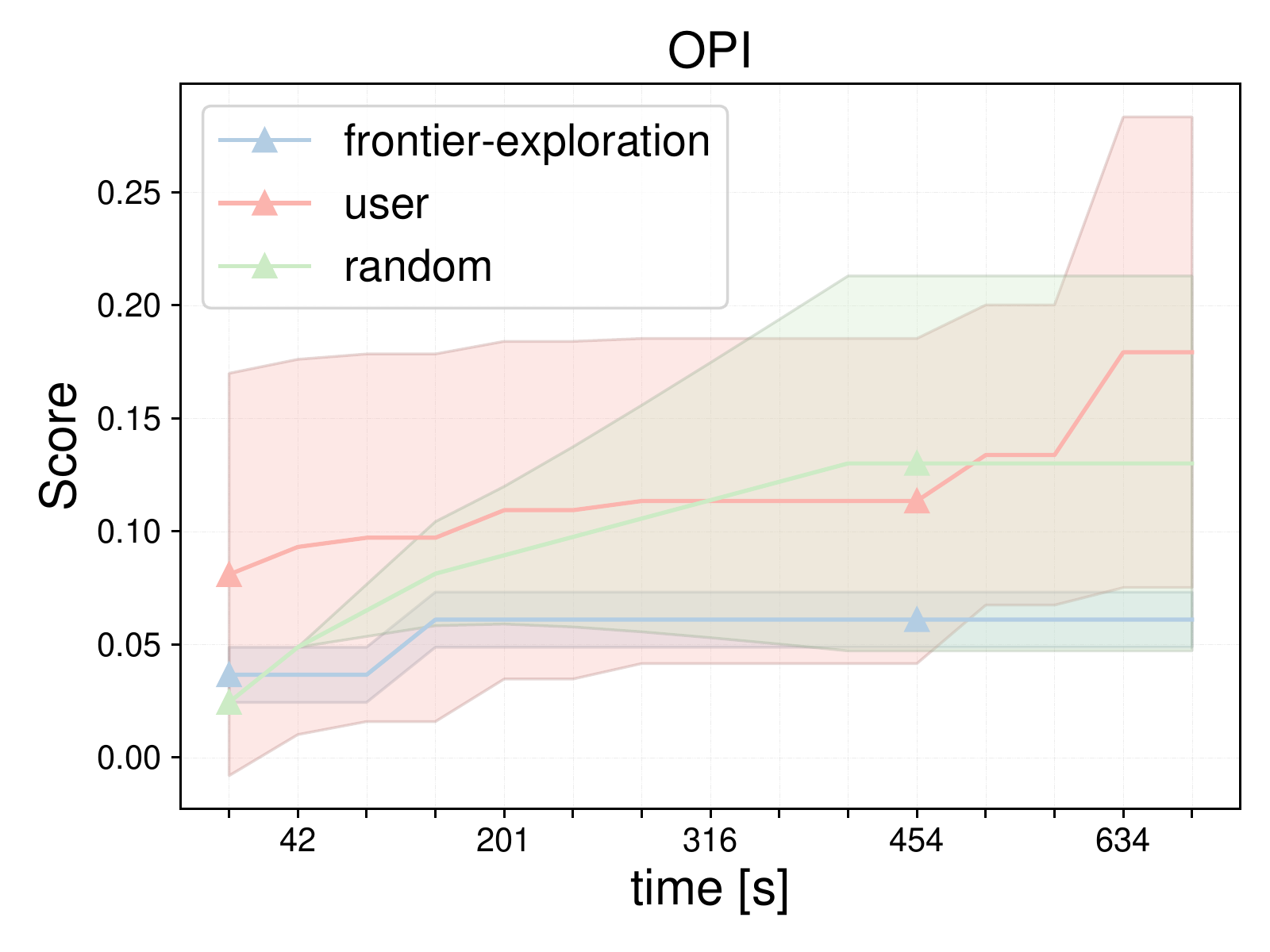}
    \label{fig:benchmarking-plots-large-office-opi}
  }
  \caption{Benchmarking metric scores in the Small Office and Large Office environments. The exploration have been conducted adopting different policies: user-driven exploration (\textit{user}); frontier-based exploration (\textit{frontier-based exploration}) and random-walk exploration (\textit{random}). On the y-axis the metric score, while on the x-axis the time of the semantic mapping session in seconds. For each algorithm, the solid line represents the average score, while the shaded area is the standard deviation computed over 5 runs of the experiment.}
  \label{fig:benchmarking-plots-2}
\end{figure}

In this section we show that given a formal representation of a semantic knowledge~\cite{Capobianco2015}, it is possible to quantitatively evaluate the performance of a semantic mapping algorithm. In particular, we evaluate semantic mapping approaches applied to robots and we measure a set of performance metrics as introduced in~\sectionname~\ref{sec:metrics}. Intuitively, these metrics support the evaluation by measuring whether the robot environment is exhaustively explored and if each object instance is thoroughly observed. In order to showcase the \rosmeery{} framework, we evaluate three different baselines and we report how they compare in building a semantic map. Nevertheless, our ultimate goal is to build a database of performance metrics of state-of-the-art semantic mapping algorithms. Baseline benchmarking sessions have been carried out in a simulated environment by using a computer with an Intel Core i7 - 6700HQ (2.6GHz) CPU, a 16GB DDR4 memory and a NVIDIA GeForce GTX 960M (4GB GDDR5 VRAM) GPU. The computer runs Ubuntu 18.04 with Gazebo 9.10 and ROS Melodic. It is worth remarking that we exploit the ROS environment as a wrapper that encapsulates \rosmeery{}, however, each node of our architecture is implemented as a standalone package and thus, independent from the specific ROS version. We execute the benchmark several times in four different environments resembling real-life scenarios. \figurename~\ref{fig:environments} shows two examples of such scenarios.

In this evaluation of the \rosmeery{} framework, we compare three different baselines to showcase how intrinsically diverse policies in exploring the environment fair against each other. To this end we implement a \textit{random}, a \textit{frontier-based} and a \textit{user-driven} exploration. This is an interesting set that highlights how a semantic map reconstruction is affected by different exploration policies. The first policy configures the robot to randomly pick a target pose within the chosen scenario and reach it.  While travelling to destination, the robot observes the environment and keeps track of the discovered objects computing the evaluation metrics. In a similar setting, the frontier-based exploration policy moves the robot by performing a coverage task, which does not take into consideration any object nor semantic entity. Finally, the user-driven exploration is considered the optional policy that attempts to optimize the semantic map, and thus the evaluation metrics. In fact, users were instructed to roam the environment by maximizing the chosen metrics. These three algorithms represent a performance indicator of a very wide spectrum of algorithms. In fact, they provide interesting insights on task-less executions (i.e. random), benchmarking sessions where the robot disregards semantic knowledge and is involved in a different task (i.e. frontier-based), and optimal policy where humans are explicitly tasked to optimize the semantic map. 

\figurename~\ref{fig:benchmarking-plots-1} and \figurename~\ref{fig:benchmarking-plots-2} collect the benchmarking metrics results over four different environments. Each environment has a different structure, topology and a varying number of objects: 35 for the \textit{Kitchen}, 21 for the \textit{Laboratory}, 11 for the \textit{Small Office}, and 41 for the \textit{Large Office} environment. All objects are common indoor elements such as, chairs, desks, cabinets, tables, fridges and other alike furniture. Noticeably, these scenarios engage the robot in an increasing -- but yet different -- set of challenges. For example, the environment in~\figurename~\ref{fig:env-3} is difficult to navigate, with several narrow passages, but it features repeated objects and topology (i.e. three parallel rows of desks). On the other hand, the environment in~\figurename~\ref{fig:env-4} is a typical kitchen and common area featuring a less predictable structure but various types of objects. To evaluate our baseline algorithms, in each of these environments, we deploy a simulated differential-drive wheeled robot equipped with a depth camera sensor and laser range-sensor.  The robot is configured to execute each of our competing algorithms (one at time) and to store metrics parameters to compute the performance indexes in accordance with the metrics introduced in~\sectionname~\ref{sec:metrics}. For each environment, we report the results in reconstructing (both without and with the confidence score), and in inferring all the objects predicates respectively.

In the different environments, we observe an expected performance of our baselines. In fact,~\figurename~\ref{fig:benchmarking-plots-1} reports a significant difference in between the user-driven baseline and the autonomous exploration policies. This is due to the fact that the Kitchen and Laboratory environments are larger and less structured and thus it becomes more difficult to locate objects for algorithms that are not tasked to do so. Conversely, as shown in~\figurename~\ref{fig:benchmarking-plots-2}, smaller and structured environments (Small and Large office), do not show a dominant approach as every object is easier to locate. In such environments, in fact, the presence of a lot of objects led to a comparable performance of the baselines. Hence, such a comparison in between different types of environments confirms that, in order to semantically map the robot scenario, a focused exploration must be provided to the robot in order to maximize the chosen metrics. Furthermore, as expected, a user-driven exploration more easily maximizes the OPI index that converges to higher values by gradually improving its performance and by completing the semantic map. On the other hand, the frontier-based and the random baselines converge to a sub-optimal value as either stop when the map is explored or simply do not observe all the environment.

\section{CONCLUSIONS}
\label{sec:conclusion}
In this work, we build upon the representation introduced in~\cite{Capobianco2015} and we propose \rosmeery{}, a reliable methodology to quantitatively evaluate environmental semantic knowledge for a robotic agent. In particular, we provide different simulated environments in \textit{Gazebo}, and measure the performance of semantic mapping algorithms with respect to two metrics: ORI and OPI. The former measures the amount of surfaces of each object observed during the robot operation, while the latter compares the number of elements in the scene and the number of instances found by the robot. Moreover, we release our evaluation pipeline as an open-source package along with different baseline algorithms. All the simulated environments, as well as the results of the user study, are free-to-download and easily executable with open software. Future directions are countless, our goal is to research specific metrics that, not only measure geometrical accuracy, but also evaluate how the robotic agent can exploit semantic knowledge to improve its actions -- as we strongly support such a point of view in benchmarking semantic maps systems. Additionally, as the several recent contributions in the design of simulated environment~\cite{di2017automatic,qiu2016UnrealCVCC,xiazamirhe2018gibsonenv}, \rosmeery{} can exploit the same approach to create more photo-realistic worlds (e.g. the Unreal Engine\footnote{The Unreal Engine is a complete open-source creation suite for game developers https://www.unrealengine.com/en-US/blog}) in order to reduce the gap between simulated and real-world deployment. Finally, we look forward to make easy-to-use interfaces to motivate new researchers to exploit our semantic mapping benchmarking framework.

\bibliographystyle{splncs04} 
\bibliography{refs}
\end{document}